\documentclass[
    iicol,
    pdflatex
]{sn-jnl}

\usepackage{adjustbox}
\usepackage{amsmath} 
\usepackage{amssymb}  
\usepackage{siunitx}
\usepackage{tensor} 
\usepackage{comment}
\usepackage{booktabs}
\usepackage{xcolor}
\usepackage{multirow}
\usepackage{url}
\usepackage{bm}
\usepackage{placeins}
\usepackage{hyperref}

\DeclareSIUnit\pixel{px}
\DeclareSIUnit\mac{MAC}
\DeclareSIUnit\rad{rad}
\DeclareSIUnit\frame{frame}
\DeclareSIUnit\null{\relax}
\usepackage[T1]{fontenc}

\newcommand{\rebuttal}[1]{\noindent#1}
\newcommand{\rebuttaltable}{}
\newadjustimage{\rebuttalfigure}[1]{width={#1}}

\jyear{2023}%

\makeatletter
\patchcmd{\ps@headings}
{\hbox to \hsize{\hfill Springer Nature 2021 \LaTeX\ template\hfill}}
{\hbox to \hsize{}}
{}
{}
\patchcmd{\ps@headings}
{\hbox to \hsize{\hfill Springer Nature 2021 \LaTeX\ template\hfill}}
{\hbox to \hsize{}}
{}
{}
\patchcmd{\ps@titlepage}
{\hbox to \hsize{\hfill Springer Nature 2021 \LaTeX\ template\hfill}}
{\hbox to \hsize{This paper has been accepted for publication in the Journal of Intelligent \& Robotic Systems. \copyright 2024 Springer.}}
{}
{}
\makeatother

\begin{document}



\title[Vision-State Fusion]{Vision-State Fusion: Improving Deep Neural Networks for Autonomous Robotics}


\author*[1]{\fnm{Elia} \sur{Cereda}}\email{elia.cereda@idsia.ch}
\author[1]{\fnm{Stefano} \sur{Bonato}}
\author[1]{\fnm{Mirko} \sur{Nava}}
\author[1]{\fnm{Alessandro} \sur{Giusti}}
\author[1,2]{\fnm{Daniele} \sur{Palossi}}

\affil[1]{
    \orgdiv{Dalle Molle Institute for Artificial Intelligence},
    \orgname{USI-SUPSI},
    \orgaddress{\city{Lugano}, \postcode{6962}, \country{Switzerland}}}

\affil[2]{
    \orgdiv{Integrated Systems Laboratory},
    \orgname{ETH Z\"urich},
    \orgaddress{\city{Z\"urich}, \postcode{8092}, \country{Switzerland}}}

\abstract{
Vision-based deep learning perception fulfills a paramount role in robotics, facilitating solutions to many challenging scenarios, such as acrobatic maneuvers of autonomous unmanned aerial vehicles (UAVs) and robot-assisted high-precision surgery.
Control-oriented end-to-end perception approaches, which directly output control variables for the robot, commonly take advantage of the robot's state estimation as an auxiliary input.
When intermediate outputs are estimated and fed to a lower-level controller, i.e. mediated approaches, the robot's state is commonly used as an input only for egocentric tasks, which estimate physical properties of the robot itself.
In this work, we propose to apply a similar approach for the first time -- to the best of our knowledge -- to non-egocentric mediated tasks, where the estimated outputs refer to an external subject.
We prove how our general methodology improves the regression performance of deep convolutional neural networks (CNNs) on a broad class of non-egocentric 3D pose estimation problems, with minimal computational cost. 
By analyzing three highly-different use cases, spanning from grasping with a robotic arm to following a human subject with a pocket-sized UAV, our results consistently improve the R\textsuperscript{2} regression metric, up to +0.51, compared to their stateless baselines.
Finally, we validate the in-field performance of a closed-loop autonomous cm-scale UAV on the human pose estimation task.
Our results show a significant reduction, i.e., 24\% on average, on the mean absolute error of our stateful CNN, compared to a State-of-the-Art stateless counterpart.
}




\maketitle


\section*{Supplementary Material}
In-field testing and demonstration video: \href{https://youtu.be/LX0seyXWQKI}{https://youtu.be/LX0seyXWQKI}.

\section{INTRODUCTION}\label{sec:intro}

Vision-based deep convolutional neural networks (CNNs) are fueling intelligent robotics, from industrial manipulators~\cite{pinto2016supersizing} to nano-sized unmanned aerial vehicles (UAVs)~\cite{frontnet} -- as big as the palm of a hand and weighting a few tens of grams.
\rebuttal{
In this work, we consider a broad class of robot perception tasks in which a robot has to estimate the relative pose of a subject of interest from high-dimensional data by an onboard sensor, such as camera images.
}

\begin{figure}[t]
	\centering
	\rebuttalfigure{\columnwidth}{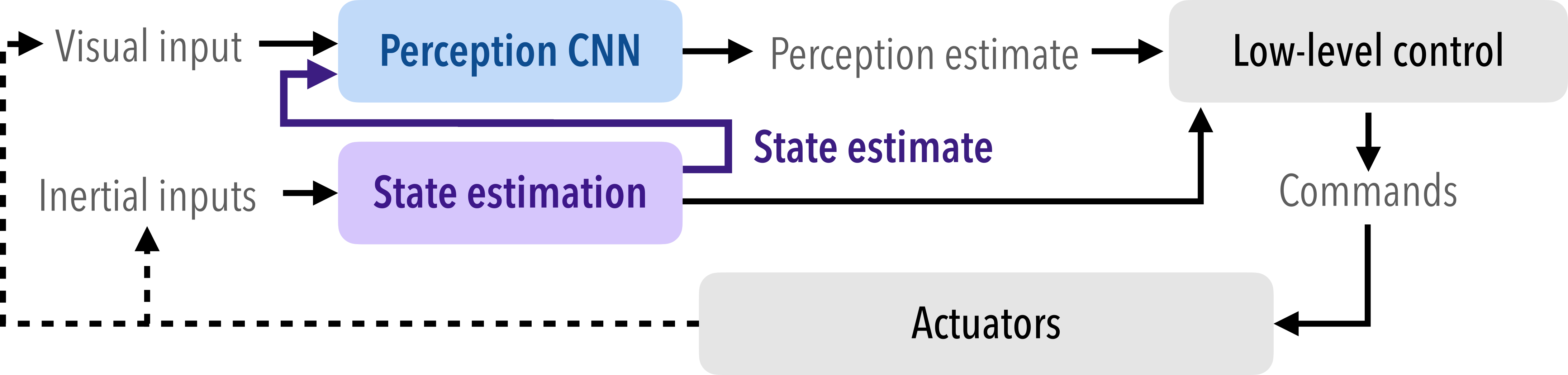}
	\caption{\rebuttal{Robotics system architecture with proposed auxiliary state input to a non-egocentric perception CNN.}}
	\label{fig:system}
\end{figure}

\rebuttal{
State estimation is a fundamental component of any robotic system, yet, not all perception tasks take advantage of this readily information about the robot’s own state.
It is common for control-oriented tasks, in which the output is fed directly into the robot's control loops (i.e., end-to-end CNNs), to use both visual data and the system's state as inputs~\cite{fast-wild,deep-acrobatics}.
Similarly, also in egocentric perception tasks, where the output refers to the system itself (e.g., egomotion, state estimation, visual odometry), it is natural to consider the system's current state to determine its future one~\cite{vinet,deepvio}.
Still, there is a class of robot perception tasks not -- yet -- vastly characterized by the use of state information as additional input: \textit{non-egocentric perception tasks}, whose output refers to subjects external to the system.
}

\FloatBarrier

\begin{figure}[t]
	\centering
	\includegraphics[width=\columnwidth]{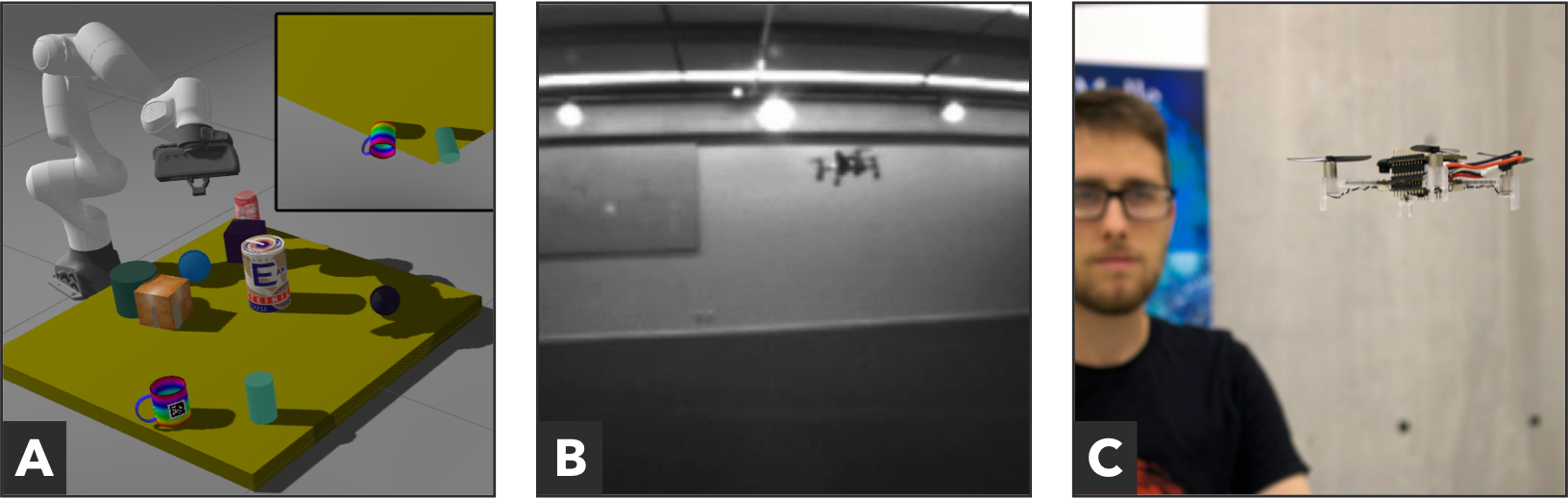}
	\caption{Our 3D pose estimation use cases: A) arm-to-object (simulation), B) drone-to-drone (on-board view), and C) drone-to-human (in-field test).}
	\label{fig:use-cases}
\end{figure}

\rebuttal{
This paper explores how the robot state can improve the spatial prediction performance of non-egocentric deep learning-based perception models, when available as an auxiliary input, as shown in Figure~\ref{fig:system}.
}
The robot state influences the perceived data predictably and can help interpret it. 
For example, images acquired by a camera mounted on the end effector of a robot arm depend on the arm's configuration, which affects the camera pose and the acquired images.
\rebuttal{
Similarly, images acquired by a front-looking camera mounted on a drone are affected by the drone's attitude (pitch and roll), which varies continuously during flight, e.g., a drone will pitch down to move forward and, in turn, see the horizon line towards the top of the image.
To ensure good performance in the field, perception CNNs should be as invariant as possible to these changes in the robot state.
}

Our approach is attractive because it matches how human perception works.
Recent neuroscience research found that vestibular information, which encodes the pose of the head, -- roughly analogous to the robot state -- contributes to the egocentric spatial representation of visual stimuli.
This has been verified in humans both by observing changes in spatial perception while electrically stimulating vestibular organs (i.e., actively corrupting state information)~\cite{abekawa2018disentangling, ferre2021vestibular}, and by studying geometric illusions (i.e., mistakes in spatial perception) in people with a malfunctioning vestibular system~\cite{clement2009mental,clement2012geometric}.

\rebuttal{
Designing a CNN architecture that fuses visual and state inputs is non-trivial.
Our main contribution is a general fusion methodology that results from comparing four alternative architectures and is supported by three different use cases, as shown in Figure~\ref{fig:use-cases}, built upon the perception task of vision-based 3D pose estimation of an object/subject.
}
In the first use case, the camera is mounted on the end effector of a manipulator's arm (arm-to-object -- A2O).
In the second one, a nano-drone's pose is estimated from a peer drone equipped with a low-quality forward-looking camera (drone-to-drone -- D2D).
In the last case, the human's pose is estimated from a nearby nano-drone (drone-to-human -- D2H). 
We address these perception problems each with a specific CNN, leveraging the respective State-of-the-Art (SoA).
We provide high variability in several key aspects, such as the robotic platform, the non-egocentric target of the pose estimation, the state information used as the auxiliary input, and the training data (simulation vs. reality), ultimately supporting the generality of our approach.

\rebuttal{
Our results demonstrate the effectiveness of the proposed approach compared to SoA stateless baseline models on all use cases, measured by $R^2$ regression score improvements up to $+0.51$ on the test sets.
For the D2H case, we strengthen our key findings by implementing, deploying, and testing both models in the field.
When running in real-time on a closed-loop autonomous nano-drone, the stateful CNN significantly outperforms its counterpart with a reduction in pose estimation error (MAE) of up to 37\%.
}


\section{RELATED WORK}\label{sec:related}

Many vision-based deep learning robotic approaches use the robot's state as one of the primary inputs to their models~\cite{fast-wild,deep-acrobatics,levine2016end,kalashnikov2018scalable,vinet,deepvio,pillai17vo}.
Exploiting state information is a well-established practice in SoA control-oriented approaches, i.e., methods that directly output actions for the robot's actuators, also known as end-to-end approaches.
\rebuttal{In the context of autonomous drones, Loquercio et al.~\cite{fast-wild,deep-acrobatics} achieve agile flight with end-to-end CNNs that leverage both visual features and the drone's linear and angular velocities, fusing the two information streams with either spatial~\cite{fast-wild} or temporal convolutions~\cite{deep-acrobatics}.}
Similarly, in the context of end-to-end robotic manipulation approaches, many consider the robot's state as an auxiliary input of a deep learning model.
For example, Levine et al.~\cite{levine2016end} train a reinforcement learning agent for manipulation tasks, combining the camera feed with the robot's state given by joints' encoder readings.
While Kalashnikov et al.~\cite{kalashnikov2018scalable} provide the model with only the height of the end-effector.

Another class of vision-based deep learning techniques tries to interpret high-dimensional perception inputs to achieve a spatial understanding of the world.
Our work belongs to this second category, which we call \textit{mediated} approaches.
These methods provide greater flexibility than end-to-end ones by decoupling perception and action in distinct algorithmic stages, e.g., pipelining multiple convolutional neural networks (CNNs).
Depending on their target outputs, mediated perception approaches can be categorized as either \textit{egocentric} or \textit{non-egocentric}.
The model's output refers to the robotic system in egocentric tasks. 
In contrast, in non-egocentric tasks, the perceptive process aims at understanding the properties of an external target, e.g., robotic manipulation of an object, pose estimation of a person, etc.

Egocentric algorithms, which estimate the robot's own trajectory, commonly rely on knowledge about their velocities and orientations, for example, by leveraging the onboard inertial measurement unit (IMU) in addition to visual information~\cite{vinet,deepvio,pillai17vo}.
VINet~\cite{vinet} models visual-inertial odometry as a sequence-to-sequence learning problem, where the camera and IMU data form two asynchronous input sequences, fused using an LSTM-based recurrent neural network to produce the sequence of egocentric poses.
DeepVIO~\cite{deepvio} extends this framework to work without ground-truth data, using stereo images during training as a source of self-supervision.
On the other hand, Pillai et al.~\cite{pillai17vo} propose a feed-forward architecture that takes the inertial information as an initial estimation of the motion and refines it using the visual feed.

To the best of our knowledge, SoA vision-based deep learning non-egocentric approaches do not exploit information about the robot's state within their perception process.
In autonomous robotics, examples include human pose estimation~\cite{frontnet,frontnet-bgrand}, tracking of peer drones~\cite{li2021drone2drone}, or gates localization to fly through them in an autonomous drone race~\cite{kaufmann2019beauty,jung2018perception}.
Similarly, approaches for robotic arms manipulation focus on identifying and localizing an object of interest to be grasped~\cite{pinto2016supersizing,tobin2017domain,zeng2017multi,nava2021uncertainty}.
Ultimately, both robotic domains solve their deep learning-based perception task by employing only visual feeds, e.g., front-looking monocular-camera~\cite{frontnet,frontnet-bgrand,li2021drone2drone,kaufmann2019beauty, jung2018perception}, eye-to-hand (i.e., fixed in space)~\cite{pinto2016supersizing,tobin2017domain,zeng2017multi} or eye-in-hand (i.e., attached to the end effector)~\cite{nava2021uncertainty} cameras.

\rebuttal{
Our work extends two well-established SoA \textit{stateless} solutions, i.e., that do not exploit state information in their deep learning-based non-egocentric perception task.
For our robotic arm manipulation use case (A2O), we leverage the CNN presented in~\cite{nava2021uncertainty}: a MobileNetV2-based model composed of 30 layers and amounting to approximately 1~million parameters.
The model takes only a 160 $\times$ \SI{120}{\pixel} RGB image from an uncalibrated monocular camera attached to the end effector and estimates the full 3D pose of an object of interest, a 7-element vector containing the object's ($x$, $y$, $z$) position and ($q_x$, $q_y$, $q_z$, $q_w$) orientation quaternion.
}

For the two nano-drone-related use cases (D2D and D2H), we take advantage of a recent lightweight CNN, PULP-Frontnet~\cite{frontnet}, capable of running up to $\sim$50 frame-per-second entirely aboard a microcontroller-based \SI{27}{\gram} drone.
Initially developed for the drone-to-human pose estimation task, we extend this CNN to the more challenging drone-to-drone scenario.
PULP-Frontnet takes as input a 160 $\times$ \SI{96}{\pixel} gray-scale image and predicts the subject's relative pose as ($x$, $y$, $z$) Cartesian position, and yaw orientation ($\phi$).
To date, very few works reach a similar level of maturity in the onboard throughput, regression performance, and energy efficiency, delivering an open-source, fully deployable, and field-tested model for vision-based pose estimation tasks aboard a nano-drone.

\rebuttal{
Finally, data augmentation is an alternative approach, orthogonal to vision-state fusion, to improve the robustness of CNN models to real-world conditions poorly represented in the train set~\cite{shorten2019augmentationsurvey,xie2020unsupervised,zheng2021spectrum}.
Task-specific augmentation strategies have also been proposed: \textit{domain randomization} to improve generalization to unseen environments~\cite{tobin2017domain,nava2021uncertainty,frontnet-bgrand}; \textit{view synthesis} to increase the density of camera poses~\cite{wan2020boosting,guerry2017snapnet,frontnet}; meta-learning approaches allow automatic tuning of task-specific augmentations~\cite{zoph2019learning}.
In our use cases, we leverage SoA augmentations, including two task-specific ones, described in Section~\ref{sec:mothodology}: domain randomization~\cite{nava2021uncertainty}, \textit{i.e.}, randomizing various aspects of the environment's appearance, in the A2O use case and pitch augmentation~\cite{frontnet}, \textit{i.e.}, synthesis of additional camera views at various pitch orientations, in D2D and D2H.
}


\section{USE CASES, MODELS, AND DEPLOYMENT} \label{sec:mothodology}

This section focuses on our non-egocentric spatial perception problems.
We start by introducing the problem of pose estimation in three different use cases: robot arm-to-object (A2O), drone-to-drone (D2D), and drone-to-human (D2H).
Then, we present the corresponding SoA CNN model used as the baseline for each use case in our comparisons.
Finally, we analyze the problem of feeding state information into a CNN model, identifying four general techniques, i.e., applicable to any CNN architecture.
We evaluate them in terms of regression performance and computation/memory cost on the D2H use case, identifying the best one, which we then apply to all three use cases.
We conclude this section by presenting our nano-drone prototype, where we deploy and field-proven our vision-state fusion implementation for the D2H use case.

\subsection{Robot arm-to-object: A2O} \label{subsec:m2o-methodology}

In this first scenario, we consider the task of visually estimating the full 3D pose of an object of interest, i.e., a rainbow-colored mug, defined with respect to the base of a robot arm.
We challenge the estimation problem by introducing multiple decoy objects sharing similar shapes with the target mug, such as cans and cylinders, as shown in Figure~\ref{fig:use-cases}-A.
We use a seven degree-of-freedom manipulator (Panda by Franka Emika), using an \textit{eye-in-hand} configuration, i.e., a downward-looking camera is attached to one side of the end-effector.


Based on the background work, we use the Gazebo simulator to collect our dataset, generating multiple environments:
each environment consists of a 90$\times$\SI{90}{\centi\meter} table, on top of which we place the object of interest along with decoy objects.
Using domain randomization~\cite{tobin2017domain}, we randomly change the pose of the object of interest; the scale, color, and pose of the decoy objects; the texture of the table; and the scene lighting direction and intensity.
Before acquiring an image of the scene from the camera, the robot is tasked to move its end-effector to a random pose within the environment, sampled from a sphere of radius \SI{50}{cm} overarching the table, and oriented to look towards a random point on the table's surface.
For each environment, the robot moves to a random pose and acquires an image 32 times, before a new environment is generated.
The robot is controlled by the MoveIt planner implementation for ROS~\cite{coleman2014reducing}.
In total, we obtain approximately 240k sample images over 160 environments -- 120 used for training, 20 for validation, and 20 for testing.

For our proposed stateful model on this use case, we consider a 7-element state vector as an additional input, representing the full 3D pose of the camera w.r.t. the robot's base in the same form as the regression output: position plus orientation quaternion.

\subsection{Drone-to-Drone: D2D} \label{subsec:d2d-methodology}

In the second use case, we consider the challenging problem of localizing a palm-sized nano-drone (i.e., 10-cm diameter) in an indoor environment, using low-resolution images acquired by a nearby peer nano-drone.

Our dataset is collected over multiple flights in a room equipped with a motion capture system, which provides millimeter-precision tracking of the drones at \SI{100}{\hertz}.
During the data collection process, two human pilots control the two peer nano-drones paying attention to maximize the variance in the acquired images and relative positions, while keeping each drone within its peer's field of view.
To maximise data collection efficiency, each drones simultaneously acts as an observer and as a target for the other drone, resulting in two parallel streams of samples each containing the camera images of one drone.
Combining the two streams, the final dataset is composed of 10k samples over 21 flights, where each sample consists of one camera image, the corresponding observer drone's onboard state, and  target drone's relative pose.
Before starting the training procedure, we filter our dataset to remove all images where the target drone is either not visible or at a distance greater than \SI{2}{m}.
This process results in 1805 samples, 60\% of which are used for training, 10\% for validation, and 30\% for testing. 
Finally, we apply different photometric augmentations to increase the number of training samples, such as gamma correction, dynamic range changes, generation of synthetic noise/blur, vignetting, and a horizontal flip.

For our stateful model on this use case, we propose to feed an additional 2-element state input to the model containing the observer drone's onboard estimate of its pitch and roll.

\begin{figure}[t]
	\centering
	\includegraphics[width=\columnwidth]{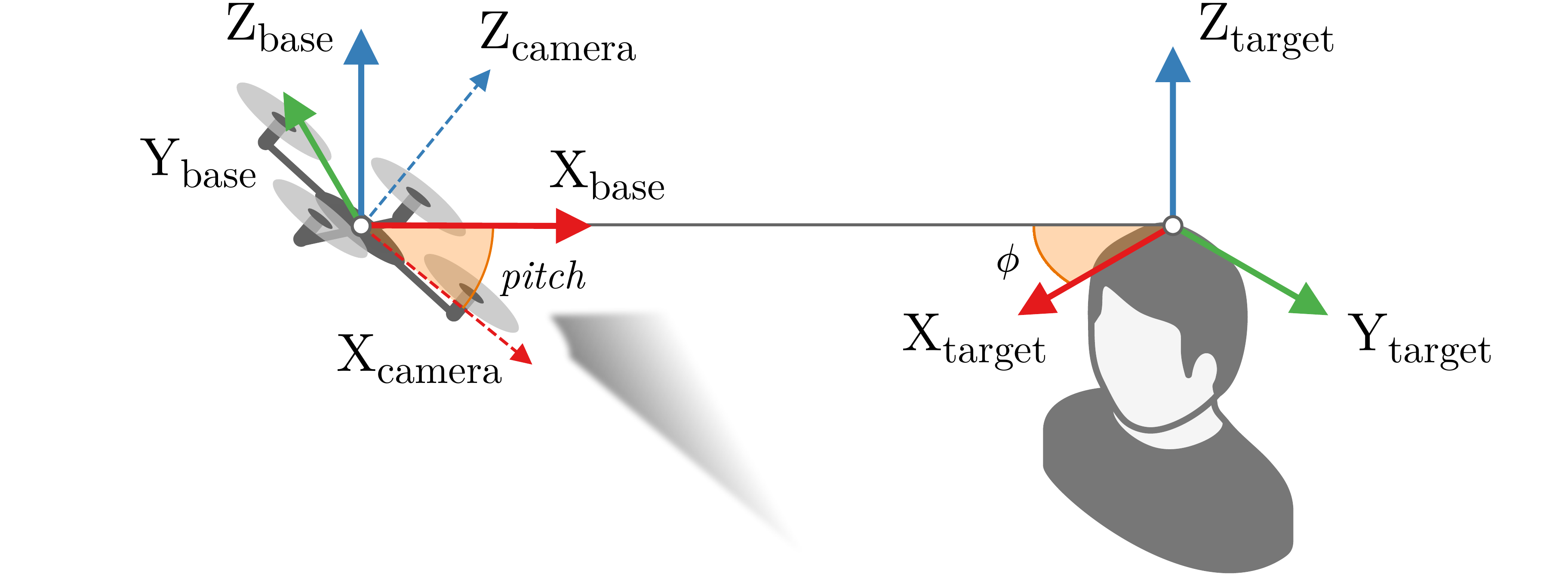}
	\caption{Reference frames in the D2H use case.}
	\label{fig:d2h-task}
\end{figure}

\subsection{Drone-To-Human: D2H} \label{subsec:d2h-methodology}

The third use case considers the task of estimating the pose of a human subject from a nano-drone flying in their vicinity.
As for the previous use case, we base our model on the SoA PULP-Frontnet architecture.
This model estimates four components of the subject's relative pose w.r.t. the observer drone's base frame: its $x$, $y$, and $z$ Cartesian position, and its yaw orientation $\phi$.
As shown in Figure~\ref{fig:d2h-task}, our models estimate the $\mathbf{target}$ relative pose w.r.t. $\mathbf{base}$, with yaw $\phi$ defined as the angle between $X_{\mathbf{target}}$ and $X_{\mathbf{base}}$.
We use a reference frame $\mathbf{base}$ identical to $\mathbf{camera}$ with $\mathit{roll}$ and $\mathit{pitch}$ angles set to zero, so that $Z_{\mathbf{base}}$ is always aligned with the world z-axis.

\begin{figure*}[htb]
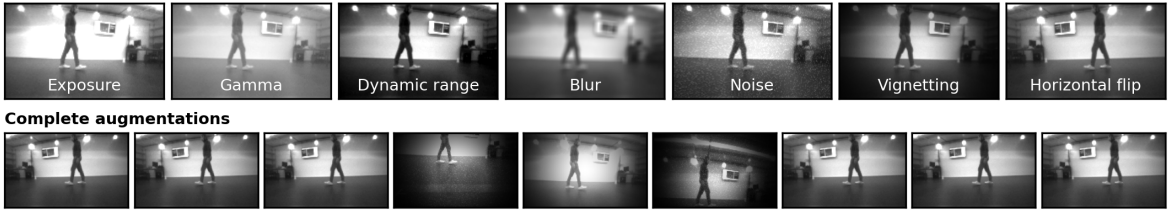

\centering
\rebuttalfigure{0.98\textwidth}{augmentations}
\caption{\rebuttal{Individual photometric data augmentations (top). Ten images produced by the full augmentation pipeline (bottom).}}
\label{fig:augmentations}
\end{figure*}

\rebuttal{
Data for our experiments consists of a combination of samples collected with the drone fixed on top of a wheeled cart (20\%) and samples collected in flight while controlled by a human pilot (80\%), in two different mocap-equipped laboratories.
In total, we collect 12k samples from 17 different human subjects of varying age, height, ethnicity, and clothing. Each sample is composed of a camera frame and associated ground-truth poses and state estimation.
Three subjects (4.7k samples) are kept for testing, while the remaining 14 subjects (7.3k samples) are split as 90\% training and 10\% validation.
}

\rebuttal{
We also synthetically increase the available training data through a number of standard photometric data augmentations that improve robustness to illumination changes -- exposure, gamma correction, dynamic range adjustments, Gaussian noise, and blurring -- followed by vignetting and horizontal flipping with 50\% probability.
We provide visual examples of each augmentation in Figure~\ref{fig:augmentations}.
We further apply synthetic pitch augmentation~\cite{frontnet} to improve the model's robustness to a broader range of pitch values, by synthesizing the image at a random synthetic pitch sampled uniformly from the $\pm$\SI{17}{\degree} range.
Each training sample is augmented offline, producing 10 augmented copies, from which we discard those samples where the subject is outside the field of view.
}

For this use case, we propose a stateful model which considers a single-element state input containing the drone's estimated pitch.

\subsection{Vision-state fusion techniques} \label{subsec:mlp-rationale}

We then analyze the problem of feeding state information into a feed-forward vision-based deep CNN, comparing four different techniques from the point of view of \textit{i}) regression performance and \textit{ii}) memory and computational costs.
If reducing the memory footprint and the number of multiply-and-accumulate (MAC) operations is desirable for any autonomous robot, it assumes paramount importance for resource-constrained platforms, such as the nano-drones we address in the D2D and D2H use cases.
We perform this investigation in the context of the D2H use case, leveraging the PULP-Frontnet~\cite{frontnet} CNN to precisely assess performance improvements and costs of each method proposed.
In this preliminary evaluation, we use the drone's pitch angle as state input (single scalar) and we focus on the output variable $z$: because of the close correlation between $z$ (relative altitude), pitch angle, and image formation, we expect this output to benefit the most from the additional state input.
In particular, when the drone flies at different $z$ altitudes, it can obtain similar images that can be disambiguated by checking its pitch orientation.
We measure regression performance in terms of the $R^2$ coefficient of determination -- as will be discussed more in depth in Section~\ref{sec:results} -- on a challenging test set built upon 5k images from 8 subjects in a never-seen-before environment.
We summarize the results of this evaluation in Table~\ref{tab:arch-comparison}, highlighting desirable characteristics in bold.
The baseline PULP-Frontnet (no state as input) requires \SI{300}{\kilo\byte} of memory, \SI{14}{\mega MAC}, and scores an $R^2=-0.58$ on the $z$ variable.

\begin{table}[t]
    \centering
    \rebuttaltable
    \scriptsize
    \resizebox{\columnwidth}{!}{
    \begin{tabular}{lllS[table-alignment-mode=format,table-number-alignment=left,table-format=+1.2]}
        \toprule
        \textbf{Architecture} &
        \textbf{Memory [\si{\byte}]} & 
        \textbf{Comput. [\si{\mac}]}& 
        \textbf{$\mathbf{R^2}$ on $\mathbf{z}$}\\
        \midrule
        Baseline~\cite{frontnet} & \SI{300.0}{\kilo\null} & \SI{14.00}{\mega\null} & -0.58 \\
        Single neuron & \textbf{\SI{300.0}{\kilo\null} [+4]} & \textbf{\SI{14.00}{\mega\null} [+4]} & 0.20 \\
        Fully connected & \SI{354.0}{\kilo\null} [+\SI{54}{\kilo\null}] & \textbf{\SI{14.05}{\mega\null} [+\SI{54}{\kilo\null}]} & 0.20 \\
        Double input & \textbf{\SI{300.8}{\kilo\null} [+800]} & \SI{17.00}{\mega\null} [+\SI{3}{\mega\null}] & 0.11 \\
        Multi-layer perc. & \textbf{\SI{300.1}{\kilo\null} [+120]} & \textbf{\SI{14.00}{\mega\null} [+104]} & {\bfseries 0.24} \\
        \bottomrule
    \end{tabular}
    }
    \caption{Comparison of CNN architectures for vision-state fusion in terms of memory, computation (multiply-and-accumulate operations) and regression performance.}
    \label{tab:arch-comparison}
\end{table}

\textbf{Single neuron.} This first method for introducing state information into our target CNN concatenates the scalar state to the existing 1920-unit activation in input to the last fully connected (FC) layer, therefore adding one single neuron.
The main advantage of this approach is the negligible cost, i.e., $+$\SI{4}{\byte} and $+$\SI{4}{MAC}.
The main drawback instead is the limited representative power as it can only model a linear mapping between the state and the regression output.
Nonetheless, this method improves the $R^2$ score on the $z$ output variable from -0.58 to 0.20.

\begin{figure*}[t]
	\centering
	\rebuttalfigure{\textwidth}{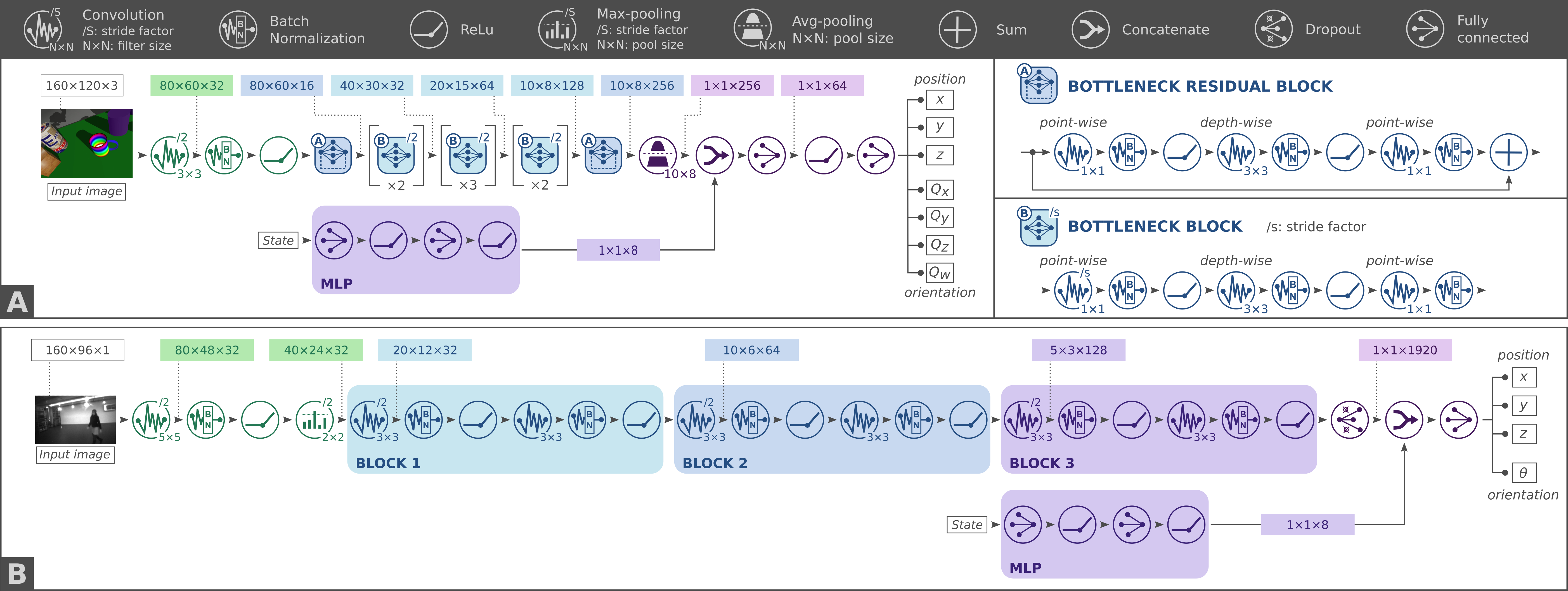}
	\caption{\rebuttal{Proposed stateful CNN architectures extended with a multi-layer perceptron branch (MLP). A) A2O use case: MobileNetV2-based CNN, with details of the repeated bottleneck residual blocks and bottleneck blocks. B) D2D and D2H use cases: PULP-Frontnet-based CNN.}}
	\label{fig:models}
\end{figure*}

\textbf{Fully connected.} To better capture non-linear mappings between the input state and the output, we consider a second approach that introduces both a second 32-unit FC layer and a ReLU non-linearity after the existing 1920-unit FC.
Despite the increased cost, i.e., $+$\SI{54}{\kilo\byte} and $+$\SI{54}{\kilo MAC}, and the additional representative power, this approach achieves a similar improvement as the previous method, i.e., $R^2=0.20$ compared to -0.58 of the stateless baseline.

\textbf{Double input.} The third strategy explored is to extend the single-channel input image by a second one where each ``pixel'' of the new channel contains the same pitch value.
This method does not require any significant modification to the original architecture and has a limited impact on the memory footprint, i.e.,  $+$\SI{800}{\byte}.
On the other hand, our experiment shows a $R^2$ score of  0.11, a smaller improvement upon the stateless baseline compared to the previous two techniques, while the computational effort grows significantly, as much as $+$\SI{3}{\mega MAC}.

\textbf{Multi-layer perceptron.} The fourth method extends the reference CNN with a small multi-layer perceptron (MLP) branch, composed of two 8-unit FC layers interleaved by ReLU non-linearities, to process the state before concatenating it to the visual features produced by the last convolutional layer.
This variant has a minimal cost both in terms of memory and computation, i.e., $+$\SI{120}{\byte} and $+$\SI{104}{MAC}, while it shows a $R^2$ score  on $z$ of 0.24, the highest among the four alternatives.
For this reason, in the rest of the work we focus on this fourth variant, i.e., the MLP, which has the best trade-off in terms of regression capability and compute/memory costs -- see Table~\ref{tab:arch-comparison}.

\subsection{Proposed CNN architectures}
We extend the SoA models of the three use cases to handle the additional state input using the MLP vision-state fusion technique identified in the previous section.
In each case, we keep the main convolutional branch of the model, unchanged, to process the camera image input and produce a vector of visual features; on the side, we introduce the proposed 2-layer MLP to process the respective state input and produce an 8-element vector of state features. The two feature vectors are then concatenated and fed to the fully connected layers.
Figure~\ref{fig:models} depicts the proposed stateful CNN architectures: the MobileNetV2-based architecture for the A2O use case and the PULP-Frontnet based architecture for the D2D and D2H use cases.

\subsection{In-field deployment: D2H} \label{subsec:in-field-setup}

To enable closed-loop in-field testing of the D2H use case, we deployed both the proposed stateful and the SoA baseline (i.e., stateless) models on the Bitcraze Crazyflie 2.1\footnote{https://www.bitcraze.io/products/crazyflie-2-1/}, a \SI{27}{\gram} nano-quadrotor.
This robotic platform exploits an STM32 microcontroller unit (MCU) to run its basic state estimation and control tasks while it is extended with the commercial AI-deck companion board~\cite{8804776}.
The AI-deck features an additional MCU: the GreenWaves Technologies GAP8, which embodies the parallel ultra-low power paradigm~\cite{pulp} through a RISC-V-based multi-core System-on-Chip (SoC).
These two processors communicate via a bidirectional UART interface.
The GAP8 is designed with two power domains: a single-core fabric controller that orchestrates the interaction with external memories/sensors and offloads computationally intensive kernels on a second 8-core cluster domain. 
The SoC's memory hierarchy relies on \SI{64}{\kilo\byte} of low-latency L1 memory shared among all cluster cores and \SI{512}{\kilo\byte} of L2 memory within the FC domain. 
The GAP8 also features two DMA engines to efficiently automate data transfers from/to all memories and external peripherals, such as the QVGA monochrome camera available on the nano-drone.
However, it provides neither data caches nor hardware floating-point units, dictating explicit data management and the adoption of integer-quantized arithmetic, respectively.

\rebuttal{
We adopt the PULP-Frontnet deployment pipeline for the CNN's main branch, composed of \SI{300}{\kilo\null} parameters and \SI{14}{\mega\mac}:
all convolutional layers are quantized to 8-bit integers and run on the GAP8's parallel cluster on images acquired by the drone's camera.
On the other hand, the proposed MLP is executed as a floating-point sequential block (soft-float), executed right after the main branch due to its minimal computational requirements, i.e., 104 multiplications (including the additional operations in the CNN's final fully connected layer).
Our soft-float choice is based on \textit{i}) the MLP's negligible workload, accounting only for 0.5\% of the total operations ($\sim20k$ cycles), and \textit{ii}) the computational overhead that would be introduced with parallel execution.
}
Finally, during the in-field mission, the current pitch is retrieved from the STM32's state estimation task and forwarded, via the UART serial interface, to the CNN running on the GAP8.

\subsection{Training and hyper-parameters}
\rebuttal{
We train all models with the Adam optimizer at a learning rate of 0.001 for 100 epochs, over the respective train sets described in Sections~\ref{subsec:m2o-methodology}--\ref{subsec:d2h-methodology}. We minimize the L1 loss between ground-truth outputs $y_i$ and model prediction $\hat{y}_i$, for each training sample $i$:
}
\begin{equation}
    L = \frac{1}{N} \sum_i \lvert y_i - \hat{y}_i \rvert
\end{equation}
\rebuttal{ 
As in previous work~\cite{frontnet,nava2021uncertainty}, each regression output of our models contributes with equal weight to the final loss function.
Further, we employ \textit{early stopping} with patience of 15 epochs and select the model that achieved the lowest validation loss.
Finally, all use cases consider a target pose defined w.r.t. the \textit{base frame} of the robot, which can always be transformed into \textit{camera frame} coordinates through a calibration procedure~\cite{clarke1998development}.
}

\rebuttal{
For in-field deployment of the D2H model, we apply 8-bit integer quantization, performing 10 epochs of quantization-aware fine-tuning~\cite{frontnet} over the train set, with Adam optimizer, learning rate $10^{-4}$, and weight decay $10^{-6}$.
}
\section{EXPERIMENTAL RESULTS}\label{sec:results}


In this section, we present four groups of experiments to assess the impact of the proposed MLP extension on the three use cases introduced in Section~\ref{sec:mothodology}.
We first investigate the regression performance of the proposed stateful models (with the MLP) comparing them against the respective SoA baseline -- called stateless models.
To further consolidate our regression analysis on the D2H scenario, we also evaluate it by deploying both models aboard a closed-loop autonomous nano-drone, i.e., relying only on onboard computation and sensors, and comparing their in-field behavior.
Finally, we discuss our key findings with respect to the SoA for all three use cases.

To guarantee the soundness of our regression analysis, for each use case, we train multiple instances of both stateful and stateless models, which differ only in the random initialization of the model parameters.
Pairs of stateful and stateless models trained from the same initial parameters share the same color in the figures below.

Our key assessment metric for regression performance is the coefficient of determination $R^2$.
It is independently computed for each target output of the CNNs, and represents a standard adimensional metric which measures the fraction of variance in the target variable explained by the model:
\begin{equation}
    R^{2} = 1 - \frac{\sum_i(y_i - \hat{y}_i)^2}{\sum_i (y_i - \bar{y})^2}
\end{equation}
where $y_i$ and $\hat{y}_i$ are, respectively, the ground-truth output and model prediction for each test sample $i$ and $\bar{y}$ the mean of ground-truth outputs.
$R^{2} = 1.0$ corresponds to a perfect regressor, while $R^{2} = 0.0$ corresponds to a dummy regressor which always predicts the mean of the test data; models can perform arbitrarily worse than this dummy regressor, leading to negative $R^{2}$ scores.

The $R^2$ score is closely related to another standard metric for regression performance, while being more conservative and easier to interpret: $R^2$ can be seen as one minus the ratio between a model's Mean Squared Error (MSE) and the dummy regressor's MSE (which corresponds to the test data's variance).
As such, a model's $R^2$ can always be interpreted in relation to the perfect and dummy regressor benchmarks. MSE, on the other hand, is expressed in the target variable's unit of measure (squared): determining whether an improvement is meaningful and comparing MSEs on different variables is only possible with the help of domain knowledge.

\begin{figure}[t]
	\centering
 	\includegraphics[width=\columnwidth]{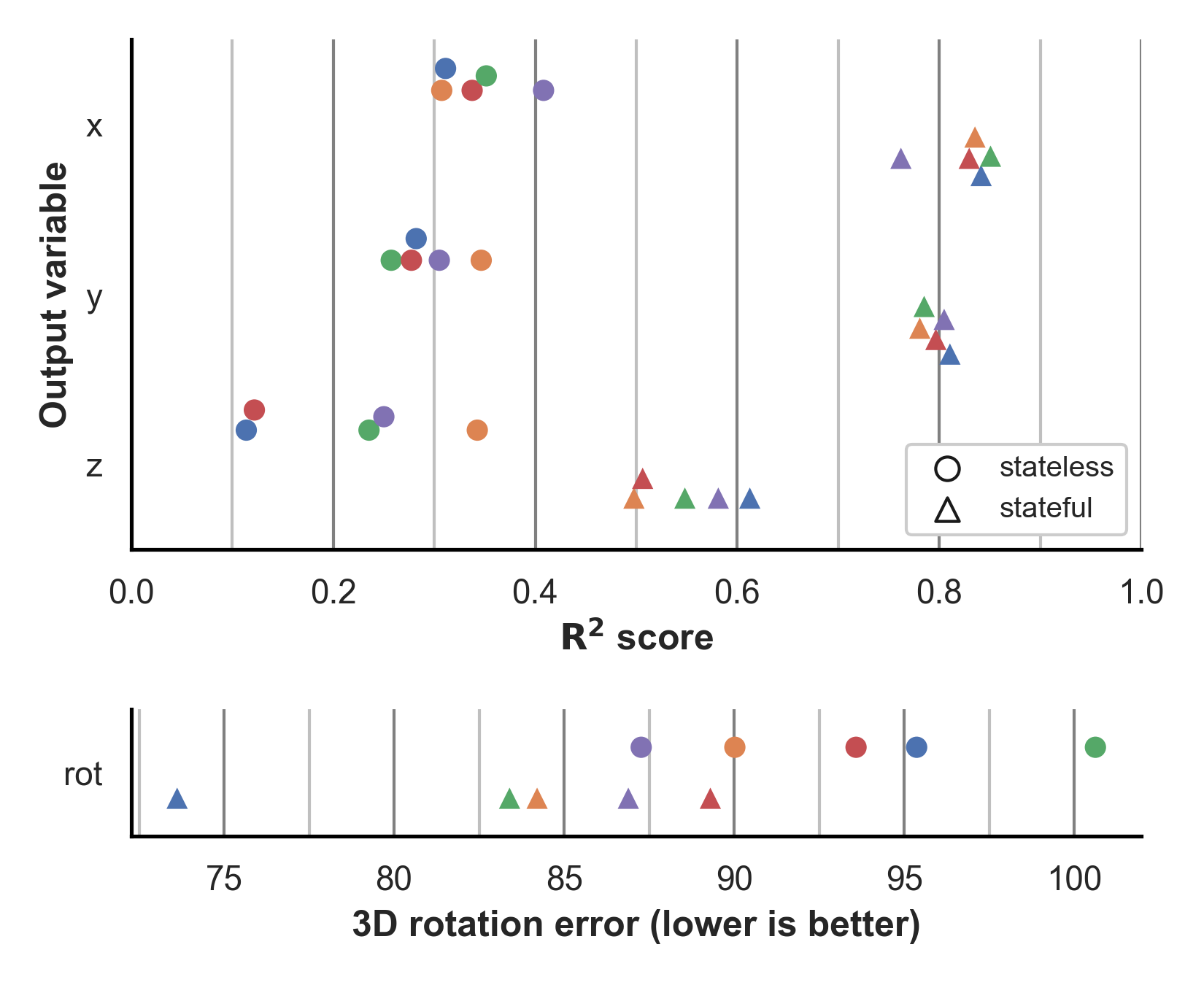}
	\caption{Test set performance comparison of $R^2$ scores on the A2O use case for both stateless and stateful models. Color identifies pairs of models trained from the same set of random initial parameters.}
	\label{fig:R2_A2C}
\end{figure}

\subsection{Regression performance: A2O} \label{sec:R2_A2C}

In Figure~\ref{fig:R2_A2C}, we compared both stateful and stateless models on the testing set, estimating the pose of the object of interest w.r.t. the robot's base frame.
We compute the models' $R^2$ score on the $x$, $y$, and $z$ components of the object's pose, as well as the 3D rotation error on the orientation component, defined as the average quaternionic distance~\cite[Eq. (4)]{mahendran20173d}.
The quaternionic distance measures the angle difference between two unit quaternions, defined between 0 and 180 degrees.
This metric is robust to the double-cover problem, i.e., it considers a quaternion $q$ and the negated one $-q$ to be at a distance 0, as opposed to more naive distance functions such as $L^p\text{-norms}$.

The median $R^2$ value of the stateful model is 0.83 on $x$ and 0.80 on $y$, which significantly improves (more than doubles) its stateless counterpart -- limited to 0.34 and 0.28 on these two variables.
Similarly, on the $z$ component, the stateful model outperforms the SoA baseline (stateless) increasing the $R^2$ from 0.24 to 0.55.
On the rotational component of the pose, the stateful model achieves a median error 10 degrees lower than the baseline model~\cite{nava2021uncertainty}.
\rebuttal{
These improvements are statistically significant to the non-parametric paired Wilcoxon test ($p \leq 0.001$ for all variables).
}
Overall, knowing the robot's state leads to significant improvements over the performance metrics, since the stateful model can directly relate the perceived images to the pose of the camera. 
This simplifies the regression task compared to its stateless counterpart, which additionally needs to estimate the camera pose from the image to correctly determine the mug's pose.

\rebuttal{
This experiment shows the value of measuring performance with $R^2$ instead of MSE:
our stateful model achieves median MSE values on $x$, $y$, and $z$ of respectively 0.020, 0.024, and 0.018, similar among the three outputs.
However, due to the varying size of the robot workspace along the three axes (90$\times$90$\times$\SI{50}{\centi\meter}), a dummy regressor would achieve MSEs of respectively 0.118, 0.116, and 0.041, which need to be accounted for when comparing the three outputs.
This is reflected in stateful R2 scores, 0.83 and 0.80 for $x$ and $y$ against only 0.55 for $z$, highlighting that our stateful model's performance on the latter output is not as good as those on the former two.
}

\begin{figure}[t]
	\centering
	\includegraphics[width=\columnwidth]{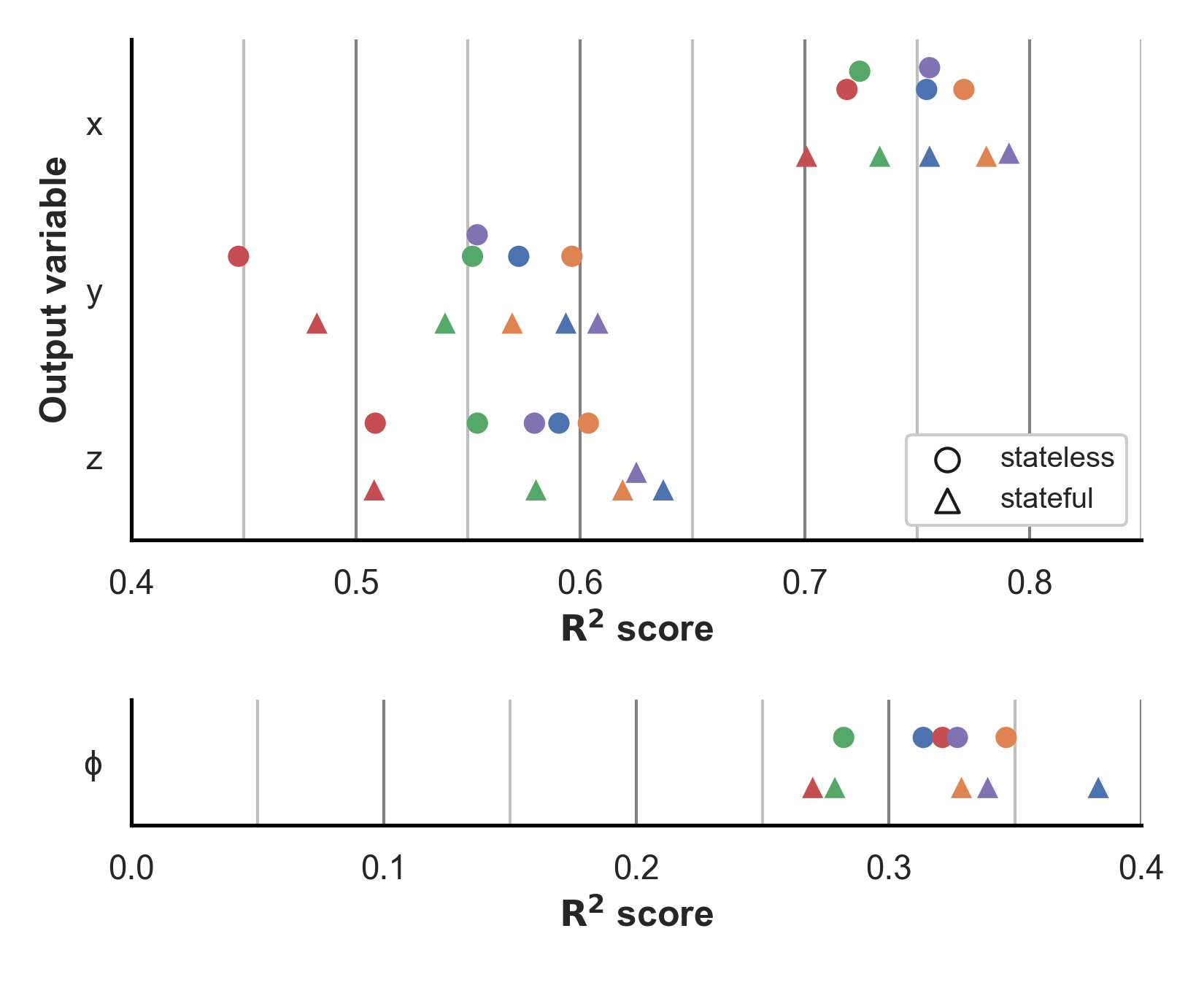}
	\caption{Test set performance comparison of $R^2$ scores on the D2D use case for both stateless and stateful models. Color identifies pairs of models trained from the same set of random initial parameters.}
	\label{fig:R2_D2D}
\end{figure}

\subsection{Regression performance: D2D} \label{sec:R2_D2D}


Figure~\ref{fig:R2_D2D} reports the $R^2$ metric for the D2D use case by analyzing both stateful and stateless models.
Both models score a similar median $R^2$ of 0.75 on the $x$ output variable, the highest among the four output variables.
A difference between models can be seen on $y$, where the stateless and stateful models score respectively 0.55 and 0.57, and on $z$, where they score respectively 0.58 and 0.62.
On the other hand, the $\phi$ variable achieves the lowest scores with a median of respectively 0.32 and 0.33, which can be ascribed to the low visibility of the \SI{10}{\cm}-diameter target nano-drone on the images, i.e., not always sufficiently big to disambiguate the drone's yaw orientation.
Considering stateful performance relative to the stateless models, Figure~\ref{fig:R2_D2D} shows an improvement on all four target variables, with median $R^2$ scores improving by $x=+0.001$, $y=+0.016$, $z=+0.039$ and $\phi=+0.008$.
\rebuttal{
This improvement is statistically significant for the $z$ variable using the non-parametric paired Wilcoxon test ($p = 0.031$), while all variables show a small positive effect.
}

These results show how $z$ is the output variable that benefits most from the pitch as input. In particular, when comparing pairs of stateless and stateful models trained from the same random initial parameters in Figure~\ref{fig:R2_D2D}, stateful models show a consistent improvement on $z$ w.r.t.~their corresponding stateless model.
This can be attributed to a strong correlation between $z$, pitch orientation, and image formation:
when the drone flies at different $z$ altitudes, it can obtain similar images which can be disambiguated by checking its pitch orientation.




Also in this case, should we measure performance exclusively with MSE,
the baseline stateless model would achieve median MSE values on $x$, $y$, $z$ and $\phi$ of respectively 0.030, 0.017, 0.007, and 0.964, which our stateful model improves to 0.030, 0.016, 0.006, and 0.953.
The dummy regressor on the same four output variables would achieve MSEs of 0.122, 0.037, 0.017, and 1.420, highlighting the difference in range between the four variables.

\begin{figure}[t]
	\centering
 	\includegraphics[width=\columnwidth]{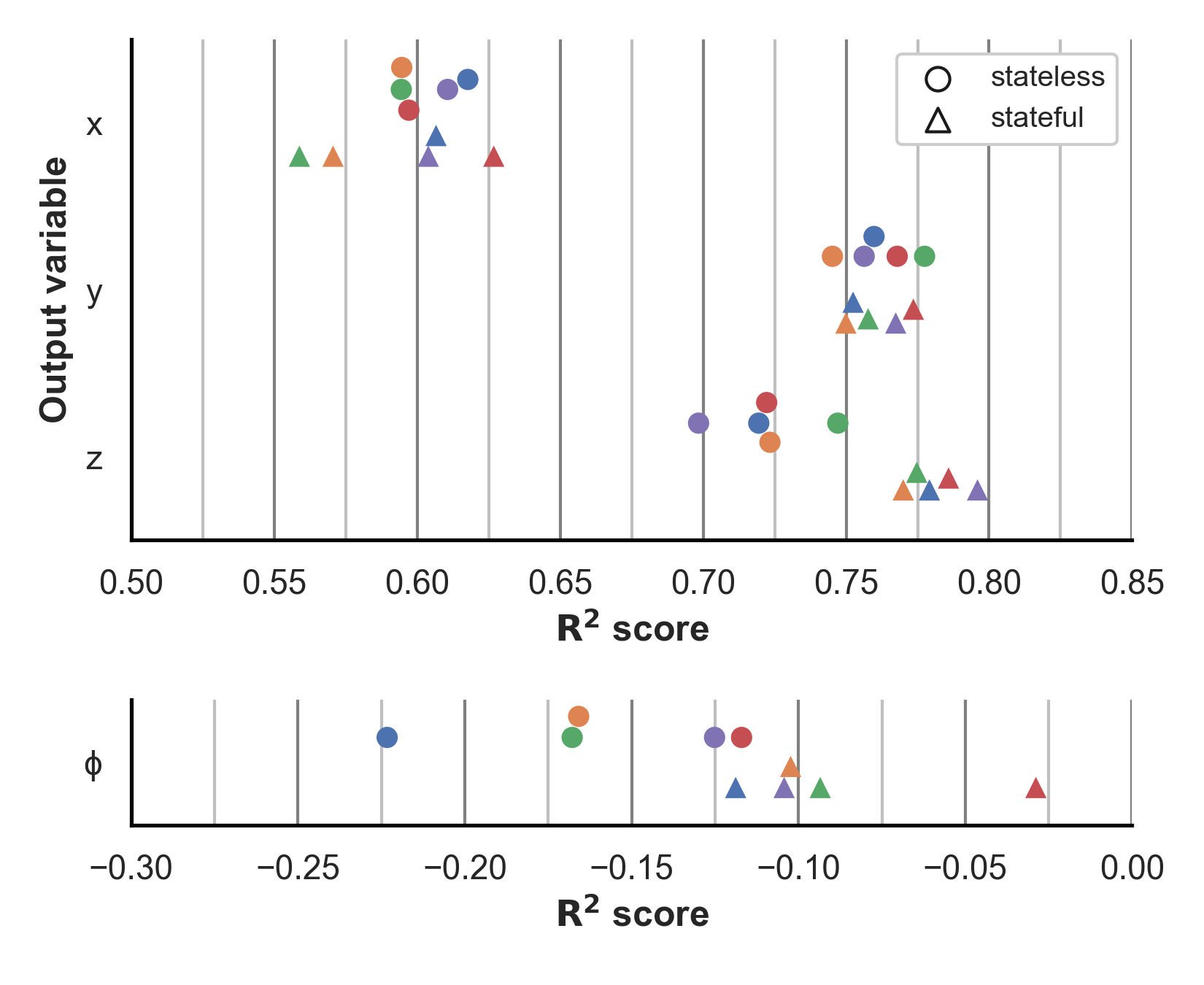}
	\caption{Test set performance comparison of $R^2$ scores on the D2H use case for both stateless and stateful models. Color identifies pairs of models trained from the same set of random initial parameters.}
	\label{fig:d2h-r2}
\end{figure}

\begin{figure*}[t]
	\centering
 	\includegraphics[width=\textwidth]{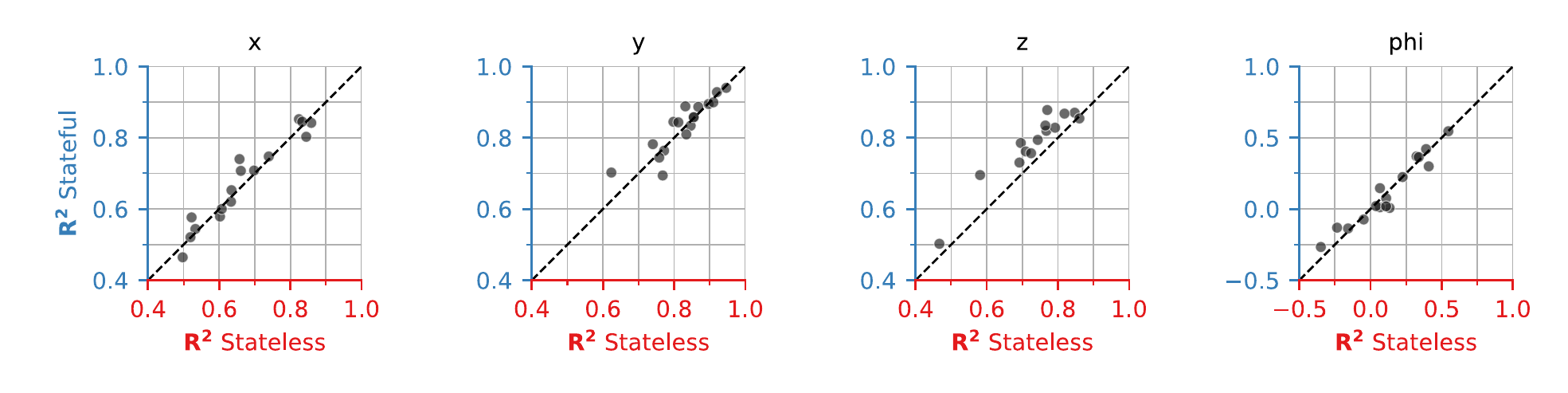}
	\caption{Regression performance in leave-one-out cross-validation of stateless and stateful models on the D2H use case.
	Each point corresponds to the pair of stateful and stateless models trained leaving out the same portion of the dataset. The dashed diagonal is the line of equivalence.}
	\label{fig:d2h-r2-crossval}
\end{figure*}


\subsection{Regression performance: D2H} \label{sec:R2_D2H}

In Figure~\ref{fig:d2h-r2}, we show the results on the third case study in terms of $R^2$ score on our test set, considering the four CNN's outputs of the relative pose between the drone and the human subject.
On the $x$ output, both models assess the same median $R^2$ of 0.60, while on $y$ and $z$, they achieve higher performance, i.e., 0.76 and 0.72 for the stateless model, 0.76 and 0.78 for the stateful one. 
Similar to the previous work~\cite{frontnet}, $\phi$ achieves lower performance compared to the other output variables, i.e., -0.17 for the stateless and -0.10 for the stateful model, while still showing improved performance by the proposed stateful one.
As a sanity check, we test our stateful model against the original PULP-Frontnet test set, which is significantly different both in terms of subjects and environments.
In this case, our model scores an $R^2$ of 0.2, doubling the performance vs. the original stateless one ($R^2$ of 0.1), due to the combination of our novel training set (2.6$\times$ larger than the previous one~\cite{frontnet}) and the proposed visual-state fusion technique.


\rebuttal{
We further analyze the regression performance of the stateful model vs. the stateless one by presenting in Figure~\ref{fig:d2h-r2-crossval} a leave-one-out cross-validation experiment.
Our \SI{12}{\kilo\null}-sample dataset, introduced in Section~\ref{subsec:d2h-methodology}, is composed of 17 subjects, which we use to train as many pairs of models (both stateful and stateless, 34 in total).
One subject is kept exclusively as test data for each pair, while the images from the remaining 16 subjects form the models' training set.
}
In Figure~\ref{fig:d2h-r2-crossval}, the horizontal axis shows the $R^2$ score of the stateless model while the vertical axis shows the $R^2$ score of the proposed stateful one; the dashed diagonal is the line of equivalence.
The median $R^2$ score on $z$ increases by $+0.051$ from the stateless to the stateful model, while the performance on $x$, $y$ and $\phi$ stays almost constant, $+0.008$, $+0.002$, and $-0.002$, respectively.
\rebuttal{
The improvement on the $z$ variable is statistically significant using the non-parametric paired Wilcoxon test ($p = 0.001$).
}
These results confirm the benefit of the stateful approach, as seen in the previous experiment, enhancing its statistical soundness due to the extensive analysis repeated on 17 pairs of models.





Also in this case, should we measure performance with MSE,
the baseline stateless model would achieve median MSE values on $x$, $y$, $z$ and $\phi$ of respectively 0.286, 0.150, 0.058, and 0.371 which our stateful model partially improves to 0.281, 0.151, 0.046, and 0.351. 
Due to the different size of the environment along the different axes, again the dummy regressor would achieve different MSEs on the same four output variables of 0.709, 0.623, 0.209, and 0.318.
By considering the $R^2$ score, we obtain a metric that can be compared among the four variables despite their different ranges.


\begin{figure*}[t]
	\centering
 	\includegraphics[width=\textwidth]{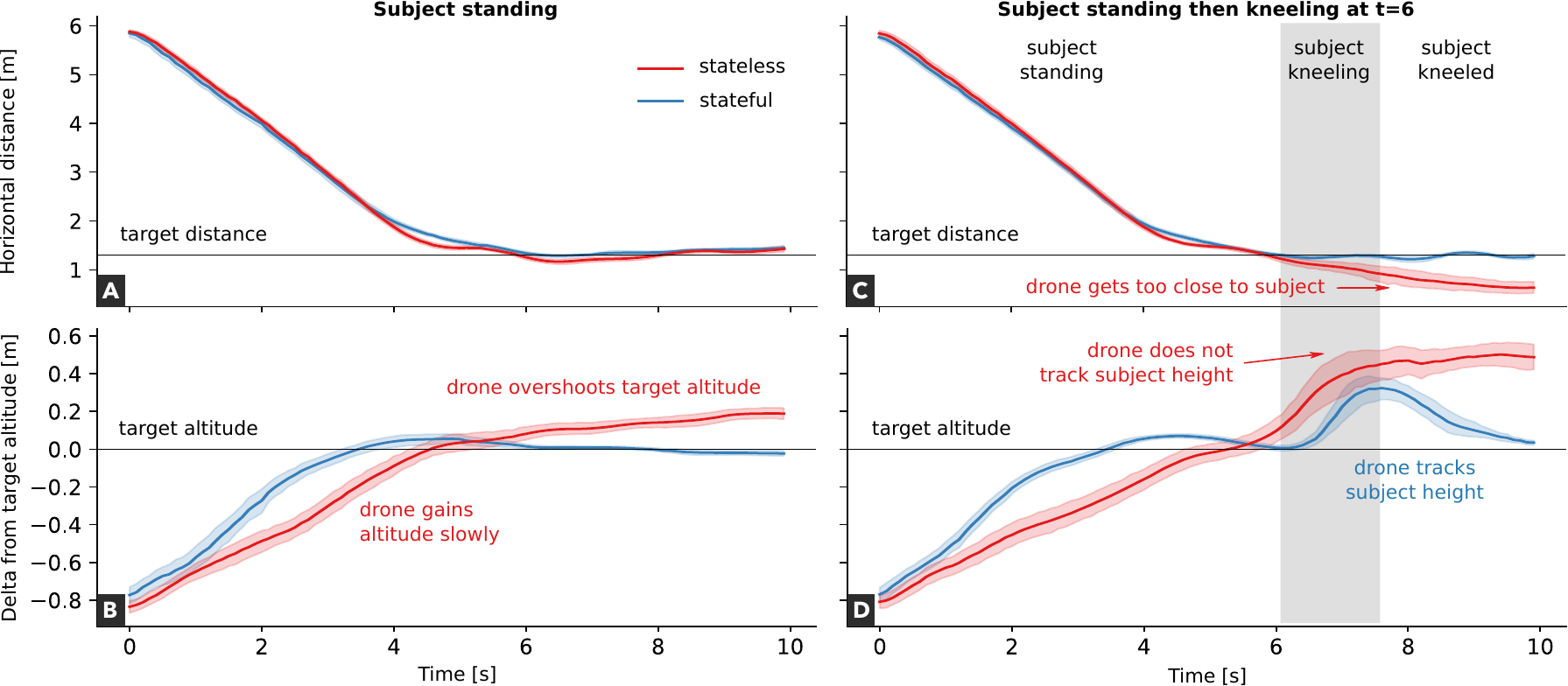}
	\caption{In-field behavior for the D2H use case, under two test scenarios: A/B) the subject standing up and remaining still for the entire test; C/D) the subject is standing still before kneeling at $t$=\SI{6}{\second}. A/C show the distance from the target position in the horizontal plane, while B/D show the distance along the vertical axis. The curves and shaded areas represent mean and $90\%$ c.i. computed from 15 flights of each model.}
	\label{fig:d2h-in-field}
\end{figure*}

\subsection{In-field experimental results: D2H} \label{sec:infield_D2H}

This section compares the stateful and stateless models with two different in-field experiments, employing a closed-loop, fully autonomous, \SI{27}{\gram} nano-drone, as described in Section~\ref{subsec:in-field-setup}.
Each of the two investigations constitutes of five $\sim$\SI{10}{\second}-long flight tests for both models.
We repeat these tests for three subjects, leading to 30 trials for each experiment -- 60 flight tests in total.
For each test, we record \textit{i}) the movements of both subject and drone, thanks to a millimeter-precise motion capture system, and \textit{ii}) the real-time model predictions from the onboard CNN.
\rebuttal{
The two models, quantized to 8-bit integers, run entirely on the GAP8 SoC aboard the nano-drone and require \SI{20.8}{\milli\second} per inference on one camera frame (a throughput of \SI{48}{\frame/\second}) at an average power consumption of \SI{96}{\milli\watt}.
}

In the first test scenario, the subject is standing still in front of the drone at $\sim$\SI{6}{\meter} distance from it.
After takeoff, the nano-drone hovers in place for a few seconds at \SI{0.5}{\meter} altitude, and then the onboard CNN's output is enabled and fed to the control loops, which are set to reach a forward target velocity of \SI{1.2}{\meter/\second}.
The desired behavior is for the drone to reach a target position at \SI{1.3}{\meter} in front of the person at eye level.
Therefore, this test requires the drone to increase its altitude during the flight towards the subject, challenging the model prediction's $x$ and $z$ components.

Figure~\ref{fig:d2h-in-field}-A/B shows the two models' behavior under this test scenario, named \textit{subject standing}.
In Figure~\ref{fig:d2h-in-field}-A, we see the drone's distance from the target in the horizontal plane.
Both models exhibit almost identical good behavior (stateless model slightly higher oscillations) by reaching the target position in around \SI{6}{\second} and converging to a steady-state hovering in front of the subject with almost no error.
In Figure~\ref{fig:d2h-in-field}-B, we see the difference (delta) between the drone altitude and target one, over time. 
The models' behavior differs noticeably; the stateless model can not reach the target altitude and overshoots, while the stateful model achieves the correct altitude and firmly converges to the desired hovering position.

With the second test scenario, named \textit{subject standing then kneeling}, we want to challenge our models (in particular the $z$ output variable) by making the subject kneel after a time of \SI{6}{\second} from the beginning of the experiment.
The experiment starts with the subject standing still, as in the previous experiment, waiting for the drone to approach, and then kneeling.
This behaviour stresses the nano-drone's $z$ prediction by making it first increase its altitude and then suddenly reducing it.
At first, the nano-drone moves forward, then it decelerates while is approaching: a condition that corresponds to a significant negative pitch.
At this point, the drone is quite close to the subject (around \SI{1.3}{\meter} in front), which means the person's downward motion corresponds to a considerable movement in image space, which might lead to losing the target from the field of view if the drone does not quickly react.

As we can see in Figure~\ref{fig:d2h-in-field}-C/D, the stateful model exhibits superior performances by correctly tracking the subject movements.
Considering the horizontal distance, Figure~\ref{fig:d2h-in-field}-C, we see the stateless model converges to the first target position in front of the standing subject ($t$=\SI{6}{\second}), but it loses track when the subject kneels.
On the other hand, the stateful model reaches the desired horizontal position and keeps it despite the subject kneeling.
Similarly, Figure~\ref{fig:d2h-in-field}-D shows how the baseline model (stateless) struggles in dynamically adjusting its attitude when the target altitude changes, resulting in an overshooting that brings the subject outside the camera's field of view.
Instead, the stateful model first precisely converges in front of the subject standing ($t$=\SI{6}{\second}).
Then, when they kneel, increasing the delta of the target altitude ($t$=\SI{6}{\second} - $t$=\SI{7.5}{\second}), the nano-drone correctly follows the movement, reaching the new target position ($t$=\SI{10}{\second}). 

Finally, to quantitatively assess the improvement of the proposed vision-state fusion technique, we compare the prediction outputs of the models w.r.t. the motion capture system ground-truth in terms of mean absolute error (MAE).
On $x$, the MAE of the stateful vs. stateless model, decreases from \SI{0.78}{\meter} to \SI{0.54}{\meter} (\SI{-30}{\percent}), on $y$ from \SI{0.45}{\meter} to \SI{0.30}{\meter} (\SI{-34}{\percent}), and on $z$ from \SI{0.54}{\meter} to \SI{0.34}{\meter} (\SI{-37}{\percent}), while $\phi$ exhibits an almost constant trend, with a MAE of \SI{0.63}{\rad} and \SI{0.65}{\rad}, respectively.
From these in-field experiments, we demonstrate how the state (pitch) as input of our CNN leads to superior regression performance vs. the SoA baseline model (stateless), significantly when the $z$ output is most challenging to predict.
We provide both models' in-field demonstration videos at \url{https://youtu.be/LX0seyXWQKI}.



\subsection{State of the art comparison and discussion} \label{sec:discussion}

We evaluated our approach on three use cases, taking state-of-the-art baselines on the respective tasks and extending them to take advantage of the robot's state with the proposed methodology.
\rebuttal{
For the robot arm use case (A2O), we built upon the MobileNetV2-based CNN from Nava et al.~\cite{nava2021uncertainty}. 
As such, we test an established CNN architecture already applied successfully to the A2O task.
Nevertheless, this model sees by far the largest improvement when provided explicit knowledge about the robot's state, ($R^2$ increases up to $+0.514$), due to the higher complexity of the considered end effector's state space (6 degrees of freedom) compared to the other two use cases (2 and 1 degrees of freedom).
}

For the two nano-drone use cases, our stateless baseline is PULP-Frontnet~\cite{frontnet}, the first CNN for human pose estimation fully deployed and field-tested on a nano-drone. 
In the drone-to-drone (D2D) use case, we adapt PULP-Frontnet to the different task of estimating a peer drone's pose.
\rebuttal{For the same task, Li et al.~\cite{li2021drone2drone} recently proposed also a YOLOv3-based~\cite{yolov3} CNN, but a direct comparison is not possible, since neither code or data has been made public.
Nonetheless, they validate their system only on a limited test set of 48 images from a single drone, with ground-truth labels acquired through a custom UWB 3D tracking system, with unspecified accuracy.
In contrast, we collect a comprehensive test set of 542 images over 21 drone flights in varying lighting and backgrounds, while acquiring precise ground-truth poses of the two drones with a mm-precise OptiTrack mocap system.}
In addition, PULP-Frontnet is larger (8 convolutional layers instead of 5 and $12\times$ as many parameters) while running faster at inference time ($30\%$ fewer multiply-accumulate operations).
Due to these considerations, PULP-Frontnet constitutes the better choice for our D2D stateless baseline.
When extended to take advantage of state information, we show an $R^2$ improvement of up to $+0.039$ compared to the baseline.

\rebuttal{
The drone-to-human (D2H) use case, on the other hand, adopts an identical task formulation as the state-of-the-art PULP-Frontnet~\cite{frontnet}, allowing a direct comparison that demonstrates a significant benefit in our approach, both in offline regression performance (increasing $R^2$ up to $+0.051$) and in closed-loop in-flight system behavior.
Compared to previous work, we introduce a larger and more comprehensive dataset (\SI{12}{\kilo\null} samples vs. \SI{4}{\kilo\null}) that covers a larger number of subjects (17 vs. 10), a wider range of drone's states (both in-flight and static, vs. static-only).
To further ensure comparability, we also test our stateful approach on the original PULP-Frontnet test set, confirming the improvements against the state-of-the art.
}

\section{CONCLUSION} \label{sec:conclusion}

In this work, we explore how the knowledge of the robot's state can be beneficial for the correct interpretation of pure-visual sensory data in many non-egocentric perception tasks.
To support the generality of our methodology, we provide three complementary robotic use cases in which we address three different instances of 3D pose estimation problems.

\rebuttal{
Extending the input of visual deep learning models with the robot's state, yields consistent improvements on spatial perception performance in our experiments, for all use cases.
The improvements scale with the complexity of the robot's state space, ranging from +0.051 in median $R^2$ on nano-drones use cases (2DoF state, pitch and roll attitude) up to +0.514 on a robot arm (6DoF state, full pose).
}
Finally, we field-proof the \textit{drone-to-human} scenario, deploying an autonomous nano-drone that assesses an average improvement of 24\% in MAE vs.~a SoA real-world baseline.

\rebuttal{
Overall, our results across the three use cases consistently show the benefits upon state-of-the-art model performance when leveraging the state input.
In the future, recurrent neural architectures are an important research direction, to further allow vision-state models to learn temporal dynamics.
Exploiting raw sensor readings, as opposed to the outputs of robot's state estimation, will also be explored.
}

\backmatter

\section*{Statements and Declarations}
\textbf{Funding:} This work was partially supported by the Secure Systems Research Center (SSRC) of the UAE Technology Innovation Institute (TII) and the Swiss National Science Foundation (SNSF) through the NCCR Robotics.

\textbf{Competing interests:} The authors have no relevant financial or non-financial interests to disclose.

\textbf{Author contributions:} All authors contributed to the study conception and design. E.C. wrote the main manuscript text and contributed the implementation and experiments for the drone-to-human use case. S.B. contributed the drone-to-drone use case. M.N. contributed the robot arm-to-object use case. All authors read and approved the final manuscript.

\textbf{Ethics approval:} This is an observational study. No ethical approval is required for this article.

\textbf{Consent to participate:} Informed consent was obtained from all individual participants included in the study.

\textbf{Consent to publish:} The authors affirm that human research participants provided informed consent for publication of the image in Figure 1.



\bibliographystyle{elsarticle-num} 
\bibliography{bibliography}






\end{document}